\def\year{2020}\relax
\documentclass[letterpaper]{article} 
\usepackage{aaai20}  
\usepackage{times}  
\usepackage{helvet} 
\usepackage{courier}  
\usepackage[hyphens]{url}  
\usepackage{graphicx} 
\usepackage{graphicx}  
\newcommand{\citet}[1]{\citeauthor{#1} \shortcite{#1}}
\newcommand{\citep}{\cite}
\usepackage{amsmath}
\usepackage{amssymb}
\usepackage{dsfont}
\usepackage{booktabs}
\usepackage{multirow}
\DeclareMathOperator*{\argmax}{arg\,max}
\usepackage{subcaption}
\title{Discovering New Intents via Constrained Deep Adaptive \\ Clustering with Cluster Refinement}

\author{
Ting-En Lin,\textsuperscript{\rm 1, 2}
Hua Xu,\textsuperscript{\rm 1,2}\thanks{Hua Xu is the corresponding author.}
Hanlei Zhang\textsuperscript{\rm 1,2,3}\\
\textsuperscript{\rm 1}State Key Laboratory of Intelligent Technology and Systems, \\ 
Department of Computer Science and Technology, Tsinghua University, Beijing 100084, China, \\
\textsuperscript{\rm 2} Beijing National Research Center for Information Science and Technology(BNRist), Beijing 100084, China\\
\textsuperscript{\rm 3} School of Computer and Information Technology, Beijing Jiaotong University, Beijing 100044, China \\
lte17@mails.tsinghua.edu.cn, xuhua@tsinghua.edu.cn, 16281181@bjtu.edu.cn
}

\begin{document}

\maketitle

\begin{abstract}
Identifying new user intents is an essential task in the dialogue system. However, it is hard to get satisfying clustering results since the definition of intents is strongly guided by prior knowledge. Existing methods incorporate prior knowledge by intensive feature engineering, which not only leads to overfitting but also makes it sensitive to the number of clusters.
In this paper, we propose constrained deep adaptive clustering with cluster refinement (CDAC+), an end-to-end clustering method that can naturally incorporate pairwise constraints as prior knowledge to guide the clustering process. 
Moreover, we refine the clusters by forcing the model to learn from the high confidence assignments. After eliminating low confidence assignments, our approach is surprisingly insensitive to the number of clusters. 
Experimental results on the three benchmark datasets show that our method can yield significant improvements over strong baselines. \footnote{The code is available at https://github.com/thuiar/CDAC-plus}
\end{abstract}

\section{Introduction}
Discovering new user intents that have not been met is an important task in the dialogue system. By grouping similar utterances into clusters, we may identify new business opportunities and decide the future direction of system development. Since most conversational data is unlabelled, an effective clustering method can help us automatically find a reasonable taxonomy and identify potential user needs. 

However, it is not as easy as we think. On the one hand, it is difficult to estimate the exact number of new intents. On the other hand, it is hard to get desired clustering results since the taxonomy of intents is usually determined by the heuristic \cite{lin-xu-2019-deep}. For example, suppose we want to partition the data according to the technical problems encountered by users, we may end up with clustering results partitioned by question types (e.g., what, how, why). 
 
Recently, this problem has attracted the attention of researchers. For example, \citet{hakkani2015clustering} use semantic parsing to decompose user utterances into graphs and prune it into subgraphs based on frequency and entropy. \citet{Padmasundari2018} combine the results of different clustering methods and sentence representations through an ensemble approach. AutoDial \cite{shi2018auto} extracts all kinds of features, such as POS tags and keywords, and then uses the hierarchical clustering method to group the sentences. \citet{haponchyk2018supervised} use predefined structured outputs to guide the clustering process. However, all of the above methods require intensive feature engineering. Besides, those methods perform representation learn and cluster assignments in a pipeline manner, which may result in poor performance. 

\begin{figure}[t!]
  \centering  
  \includegraphics[width=0.95\columnwidth ]{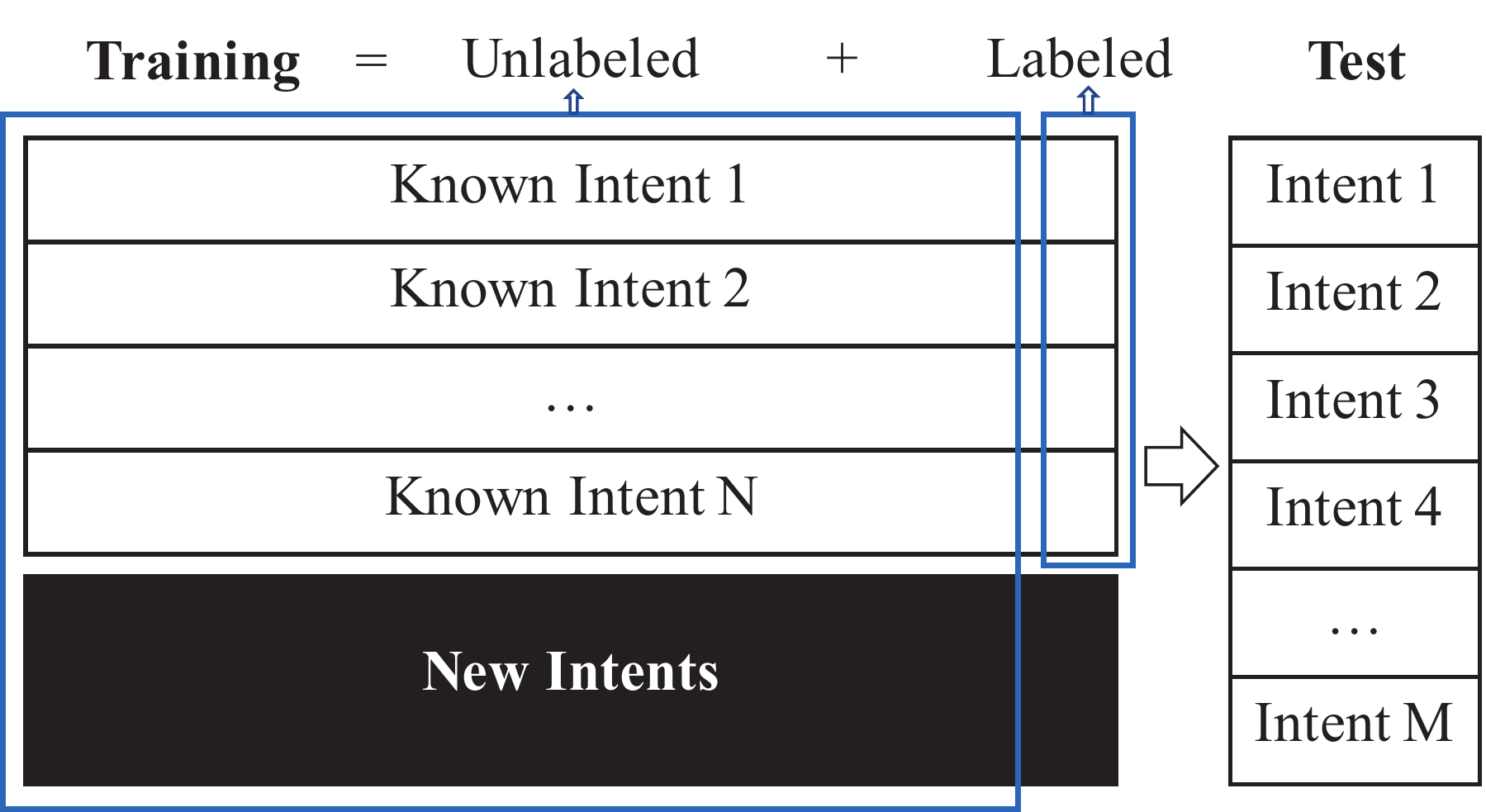}
  \caption{\label{example} An example of new intent discovery. Our goal is to find out the underlying new intents by utilizing the limited labeled data to guide the clustering process.}
\end{figure}

\begin{figure*}[t!]
  \centering  
  \includegraphics[width=2 \columnwidth ]{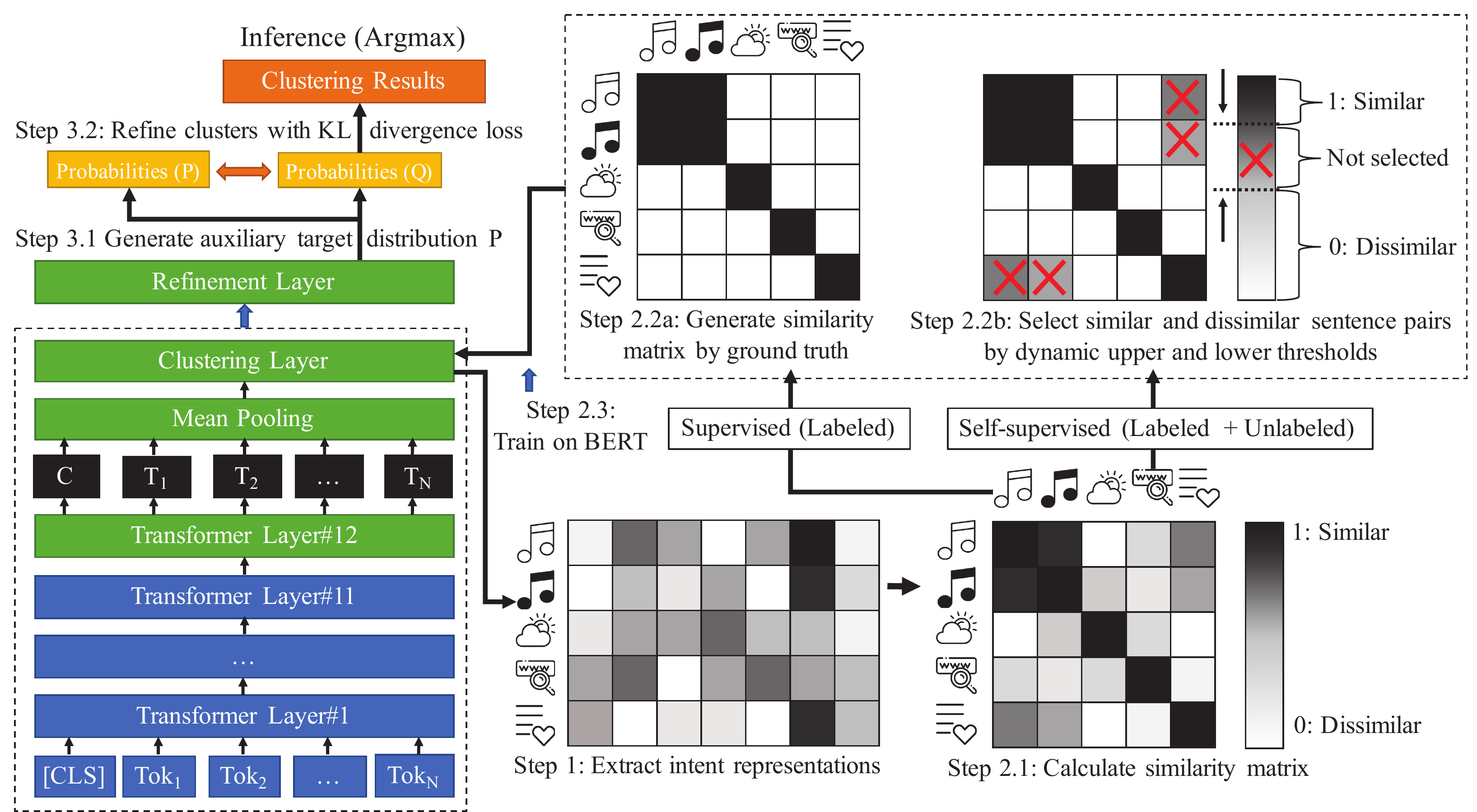}
  \caption{\label{model} The model architecture of CDAC+. We repeat step 1 and 2 iteratively until the upper and lower thresholds overlap. Then we go to step 3 to refine the clustering results further. The figure is best viewed in color. We use blue blocks to represent frozen network parameters.}
\end{figure*}

In reality, as shown in Figure \ref{example}, we may have limited labeled data and a vast amount of unlabeled data, and we do not know all the intent categories in advance. Besides, the training data is noisy because unlabeled data contain both known and unknown intents. The key is to take advantage of labeled data to improve clustering performance effectively. 

To address these issues, we propose an end-to-end clustering method that optimizes the intent representation within the clustering process. Also, we leverage the pre-trained language model, BERT \citep{devlin2018bert}, and the labeled data to aid the clustering process. As shown in Figure \ref{model}, we divide our method into three steps. First, we obtain intent representations from BERT. Second, we construct a pairwise-classification task as the surrogate for clustering by determining whether the sentence pair is similar or not. We use intent representations to calculate the similarity matrix of sentence pairs. Then, we train the network with similar or dissimilar labels, which are generated by either labeled data or dynamic similarity thresholds. We treat the pairwise constraints provided by labeled data as prior knowledge and use it to guide the clustering process. Finally, we use the auxiliary target distribution and Kullback-Leibler divergence (KLD) loss to encourage the model to learn from the high confidence assignments. We refine the intent representation and cluster assignment jointly in an end-to-end fashion. By eliminating low confidence assignments, our approach is insensitive to the number of clusters.

We summarize our contribution as follows. First, we propose an end-to-end clustering method that does not require intensive feature engineering and is insensitive to the number of clusters. Second, we demonstrate how to leverage the pre-trained language model and limited labeled data to aid the clustering process. Finally, extensive experiments conducted on three datasets show that our method can yield significant improvements compared with strong baselines.

\section{Related Work}
\subsection{Transfer Learning} 
Transfer learning uses knowledge in the source domain to help the learning process in the target domain. Types of transferred knowledge include training instance, feature representation, model parameter, and the relation among data \cite{Pan2010}. \citet{hsu2018learning} propose to transfer the pairwise similarity for cross-domain clustering. However, an extra similarity prediction model must be trained in advance. Our method transfers the relations among data through pairwise similarity and model parameters of the pre-trained language model \cite{devlin2018bert}, but does not require an extra similarity prediction model. We use the transferred knowledge to guide the clustering process. 

\subsubsection{Few-shot learning}
Few-shot learning also requires knowledge transfer from existing classes to the new classes. It focuses on classification problems, and one of the most popular methods is to transfer knowledge via the clustering approach \cite{snell2017prototypical}. Besides, its test set contains only new classes.

In contrast, our work focuses on clustering problems, and we transfer knowledge via classification approach. When we try to discover new intents, the test set usually contains both known and unknown classes, and the samples could be easily misclassified as known classes. Our setting is not only closer to reality but also more challenging.

\subsection{Unsupervised Clustering}
In the literature, various algorithms can be used for intent clustering such as K-means \cite{macqueen1967some} and agglomerative clustering \cite{gowda1978agglomerative}. However, these traditional methods are ineffective in high-dimensional data due to the limitations of feature space and the choice of predefined distance metrics. 

We can resolve this problem by using neural networks to optimize the feature space in advance. For example, STCC \cite{xu-etal-2015-short} construct a self-taught objective to learn the compressed representation and then perform K-means on it. With the development of deep learning, researchers start studying how to use the neural network to learn feature representations and cluster assignments simultaneously.

\subsubsection{Deep Neural Network-based Clustering} 
Clustering with deep neural networks is an emerging topic, and deep embedding clustering (DEC) \cite{xie2016unsupervised} opens up the possibility for it. DEC compress the TF-IDF of documents into low-dimensional representations through a stacked autoencoder. Then, they iteratively optimize the clustering objective with a self-training target distribution by KLD loss. Deep clustering network (DCN) \cite{yang2017towards} follow the idea of DEC and add penalty term on reconstruction during the process of optimizing the clustering objective. However, these methods merely compress representations and unable to capture the context effectively.

\citet{chang2017deep} propose deep adaptive clustering (DAC), which uses a pairwise-classification framework and recasts image clustering into a binary classification problem. They use convolutional neural network (CNN) to determine whether the sentence pair is similar or not. Then, they perform adaptive clustering in a self-supervised manner. The key is that the filters of CNN can naturally provide discriminative power even they are randomly initialized \cite{caron2018deep}, so the similarity between samples can be measured. However, the assumption does not work for text data. Instead, we replace CNN with the pre-trained language model and use it to measure the similarity between samples. 

\subsection{Constrained Clustering} 
Constrained clustering uses a small amount of labeled data to aid the clustering process. A paradigm is to modify the clustering objective function to satisfy the pairwise constraints. For example, COP-KMeans \cite{Wagstaff2001} use \textit{must-link} and \textit{cannot-link} between samples as hard constraints. PCK-Means \cite{basu2004active} introduce the soft constraints by allowing the constraints to be violated with violation cost. Based on PCK-Means, MPCK-Means \cite{bilenko2004integrating} use the constraints to optimize the distance metric simultaneously. \citet{wang-etal-2016-semi} extend the idea to neural networks with instance-level constraints. \citet{hsu2018learning} use an extra similarity prediction model to incorporate pairwise constraints into the clustering process. We use pairwise constraints for optimizing clustering objective and metric learning in our model.

\begin{table*}[t!]
\centering
\begin{tabular}{@{} ccccccc @{}}
\toprule
  Dataset & \#Classes (Known + Unknown) & \#Training & \#Validation & \#Test & Vocabulary & Length (max / mean) \\
\midrule
  SNIPS & 7 (5 + 2) & 13,084 & 700 & 700 & 11,971 & 35 / 9.03 \\
  DBPedia & 14 (11 + 3) & 12,600 & 700 & 700 & 45,077 & 54 / 29.97\\
  StackOverflow & 20 (15 + 5) & 18,000 & 1,000 & 1,000 & 17,182 & 41 / 9.18 \\
\bottomrule
\end{tabular}
\caption{ \label{data-stat-table}  Statistics of SNIPS, DBPedia, and StackOverflow dataset. \# indicates the total number of sentences. In each run of the experiment, we randomly select 25\% intents as unknown. Taking SNIPS dataset as an example, we randomly select 2 intents as unknown and treat the remaining 5 intents as known. }
\end{table*}

\section{Constrained Deep Adaptive Clustering \\ with Cluster Refinement}
We divide the proposed method into three steps: intent representation, pairwise-classification, and cluster refinement. The model architecture is shown in Figure \ref{model}.
\subsection{Intent Representation}
First, we use the pre-trained BERT language model to obtain intent representations. Given the $i^{th}$ sentence $x_i$ in the corpus, we take all token embeddings $[C, T_1, \cdots, T_N] \in \mathds R^{(N+1) \times H}$ in the last hidden layer of BERT and apply mean-pooling on it to get the average representation $e_i \in \mathds R^{H}$:
\begin{align}
    e_i = \text{mean-pooling}([C, T_1, \cdots, T_N]) 
\end{align}
where N is the sequence length and H is the hidden layer size. Then, we feed $e_i$ to clustering layer $g$ and obtain intent representation $I_i \in \mathds R^{k}$:
\begin{align}
    g(e_i) = I_i = W_2 ( \text{Dropout}(\text{tanh}(W_1 e_i)))
\end{align}
where $ W_1 \in \mathds R^{H \times H}$ and  $W_2 \in \mathds R^{H \times k}$ are learnable parameters, and $k$ is the number of clusters. We use clustering layer to group the high-level features and extract intent representation $I_i$ for next steps.

\subsection{Pairwise-Classification with Similarity Loss}
The essence of clustering is to measure the similarity between samples (\citeauthor{haponchyk2018supervised} \citeyear{haponchyk2018supervised}; \citeauthor{poddar2019train} \citeyear{poddar2019train}). Inspire by DAC \cite{chang2017deep}, we reframe the clustering problem as a pairwise-classification task. By determining whether the sentence pair is similar or not, our model can learn clustering-friendly intent representation. We use intent representation $I$ to compute the similarity matrix $S$:
\begin{align}
    S_{ij} = \frac{I_i I_j^T}{\| I_i \parallel \| I_j \parallel}
\end{align}
where $\|\cdot\|$ is L2 norm and $i, j \in \{1, \dots , n\}$. We denote batch size as $n$. $S_{ij}$ indicates the similarity the between sentence ${x_i}$ and $x_j$. Then, we iteratively go through supervised and self-supervised step to optimize the model.

\subsubsection{Supervised Step} Given a small amount of labeled data, we can construct the label matrix $R$:
\begin{align}
  R_{ij} :=\begin{cases}
               1, \text{ if} \quad y_{i} = y_{j}, \\
               0, \text{ if} \quad y_{i} \neq y_{j}
            \end{cases}
\end{align}
where $i, j \in \{1, \dots, n \}$. Then, we use the similarity matrix $S$ and the label matrix $R$ to compute the similarity loss  $\mathcal{L_{\text{sim}}}$: 
\begin{equation}
\begin{aligned}
    \mathcal{L}_{\text{sim}}(R_{ij}, S_{ij}  ) = -& R_{ij} \log(S_{ij}) \\
     - &(1- R_{ij}) \log(1-S_{ij}).
\end{aligned}
\end{equation}
Here we treat labeled data as priori knowledge and use it to guide the clustering process. It implies how the model should partition the data. 

\subsubsection{Self-supervised Step} First, by applying dynamic thresholds on similarity matrix $S$, we get the self-labeled matrix $\hat R$:
\begin{align}
 \hat R_{ij} :=\begin{cases}
               1, \text{if} \quad S_{ij}  > u(\lambda) \quad \text{or} \quad y_i = y_j,  \\
               0, \text{if} \quad S_{ij}  < l(\lambda) \quad \text{or} \quad y_i \neq y_j,  \\
               \text{Not selected , otherwise}
            \end{cases}
\end{align}
where $i, j \in \{1, \dots, n \}$. The dynamic upper threshold $u(\lambda)$ and the dynamic lower threshold $l(\lambda)$ are used to determine whether the sentence pair is similar or dissimilar. Note that the sentence pairs with similarities between $u(\lambda)$ and $l(\lambda)$ do not participate in the training process. In this step, we mix labeled and unlabeled data to train the model. The labeled data can provide ground truth for noisy self-labeled matrix $\hat S$ and reduce the error.

Second, we add $u(\lambda)-l(\lambda)$ as the penalty term for the number of samples. 
\begin{align}
    \min_\lambda \mathbf{E}(\lambda) = u(\lambda) - l(\lambda)
\end{align}
where $\lambda$ is an adaptive parameter that controls the sample selection, and we iteratively update the value of $\lambda$ with:
\begin{align}
    \lambda := \lambda - \eta \cdot \frac{\partial \mathbf{E}(\lambda)}{\partial \lambda}
\end{align}
where $\eta$ denotes the learning rate of $\lambda$. Since $u(\lambda) \propto -\lambda$ and $l(\lambda) \propto \lambda$, we can gradually increase $\lambda$ during the training process to decrease $u(\lambda)$ and increase $l(\lambda)$. It allows us to gradually select more sentence pairs to participate in the training process. It may also introduce more noise to $\hat R$.

Finally, we use the similarity matrix $S$ and the self-labeled matrix $\hat R$ to compute the similarity loss  $\mathcal{\hat L_{\text{sim}}}$: 
\begin{equation}
\begin{aligned}
    \mathcal{\hat L}_{\text{sim}}(\hat R_{ij}, S_{ij}  ) = -& \hat R_{ij} \log(S_{ij}) \\
     - &(1- \hat R_{ij}) \log(1-S_{ij}).
\end{aligned}
\end{equation}
As the thresholds change, we train the model from easily classified sentences pair to hardly classified sentences pair iteratively to obtain the clustering-friendly representation. When $u(\lambda) \leq l(\lambda)$, we stop the iterative process and move to the refinement stage.

\subsection{Cluster Refinement with KLD loss}
We adopt the idea of \citet{xie2016unsupervised} and refine the cluster assignments via an expectation-maximization approach iteratively. The intuition is to encourage the model to learn from the high confidence assignments. First, given the initialized cluster centroids $U  \in \mathds R^{k \times k}$ saved in the refinement layer, we calculate the soft assignment between intent representations and cluster centroids. Specifically, we use Student’s t-distribution as a kernel to estimate the similarity between intent representation $I_i$ and cluster centroid $U_j$:
\begin{align}
    Q_{ij} = \frac{(1+ \parallel I_i - U_j \parallel ^2)^{-1}}{\sum_{j'}(1+ \parallel I_i - U_j \parallel ^2)^{-1}} 
\end{align}
where $Q_{ij}$ represents the probability (soft assignment) that the sample $i$ belongs to the cluster $j$. Second, we use the auxiliary target distribution $P$ to force the model to learn from the high confidence assignments, thereby refining the model parameters and cluster centroids. We define target distribution $P$ as follows:
\begin{align}
    P_{ij} = \frac{Q^2_{ij}/f_i}{\sum_{j'} Q^2_{ij'}/f_{j'}}
\end{align}
where $f_i = \sum_i Q_{ij}$ denotes the soft cluster frequencies. Finally, we minimize the KLD loss between $P$ and $Q$:
\begin{align}
    \mathcal{L}_{\text{KLD}} = KL(P\|Q) = \sum_i \sum_j P_{ij} \log \frac{P_{ij}}{Q_{ij}} 
\end{align}
Then, we repeat the above two steps until the cluster assignment changes less than $\delta_{label} \%$ in two consecutive iterations. Finally, we inference cluster $c_i$ results as follows:
\begin{align}
    c_i = \argmax_{k} Q_{ik}
\end{align}
where $c_i$ is the cluster assignment for sentence $x_i$.

\section{Experiments}
\subsection{Datasets}
We conduct experiments on three publicly available short text datasets. The detailed statistics are shown in Table \ref{data-stat-table}.

\begin{table*}[t!]
\centering
\begin{tabular}{@{\extracolsep{4pt}}ccccccccccc}
\toprule
\centering
 &  & \multicolumn{3}{c}{SNIPS} & \multicolumn{3}{c}{DBPedia} & \multicolumn{3}{c}{StackOverflow} \\
 \addlinespace[0.1cm] \cline{3-5} \cline{6-8} \cline{9-11}  \addlinespace[0.1cm]
 & Method & NMI & ARI & ACC & NMI & ARI & ACC & NMI & ARI & ACC \\
\midrule
\multirow{6}{*}{Unsup.} 
& KM & 71.42 & 67.62 & 84.36 & 67.26 & 49.93 & 61.00 & 8.24 & 1.46 & 13.55 \\
& AG & 71.03 & 58.52 & 75.54 & 65.63 & 43.92 & 56.07 & 10.62 & 2.12 & 14.66 \\
& SAE-KM & 78.24 & 74.66 & 87.88 & 59.70 & 31.72 & 50.29 & 32.62 & 17.07 & 34.44 \\
& DEC & 84.62 & 82.32 & 91.59 & 53.36 & 29.43 & 39.60 & 10.88 & 3.76 & 13.09 \\
& DCN & 58.64 & 42.81 & 57.45 & 54.54 & 32.31 & 47.48 & 31.09 & 15.45 & 34.26 \\
& DAC & 79.97 & 69.17 & 76.29 & 75.37 & 56.30 & 63.96 & 14.71 & 2.76 & 16.30 \\
& BERT-KM & 52.11 & 43.73 & 70.29 & 60.87 & 26.6 & 36.14 & 12.98 & 0.51 & 13.9 \\
\midrule
\multirow{4}{*}{Semi-sup.} 
& PCK-means & 74.85 & 71.87 & 86.92 & 79.76 & 71.27 & 83.11 & 17.26 & 5.35 & 24.16 \\
& BERT-KCL & 75.16 & 61.90 & 63.88 & 83.16 & 61.03 & 60.62 & 8.84 & 7.81 & 13.94 \\
& BERT-Semi & 75.95 & 69.08 & 78.00 & 86.35 & 72.49 & 75.31 & 65.07 & 47.48 & 65.28 \\
& CDAC+ & \textbf{89.30} & \textbf{86.82} & \textbf{93.63} & \textbf{94.74} & \textbf{89.41} & \textbf{91.66} & \textbf{69.84} & \textbf{52.59} & \textbf{73.48} \\
\bottomrule
\end{tabular}
\caption{ \label{results-main}  
The clustering results on three datasets. We evaluate both unsupervised and semi-supervised methods.  
}
\end{table*}

\begin{table*}[t!]
\centering
\begin{tabular}{@{\extracolsep{4pt}}ccccccccccc}
\toprule
\centering
 &  & \multicolumn{3}{c}{SNIPS} & \multicolumn{3}{c}{DBPedia} & \multicolumn{3}{c}{StackOverflow} \\
 \addlinespace[0.1cm] \cline{3-5} \cline{6-8} \cline{9-11}  \addlinespace[0.1cm]
 & Method & NMI & ARI & ACC & NMI & ARI & ACC & NMI & ARI & ACC \\
\midrule
\multirow{3}{*}{Unsup.} 
& DAC & 79.97 & 69.17 & 76.29 & 75.37 & 56.30 & 63.96 & 14.71 & 2.76 & 16.30 \\
& DAC-KM & 86.29 & 82.58 & 91.27 & 84.79 & 74.46 & 82.14 & 20.28 & 7.09 & 23.69 \\
& DAC+ & 86.90 & 83.15 & 91.41 & 86.03 & 75.99 & 82.88 & 20.26 & 7.10 & 23.69 \\
\midrule
\multirow{3}{*}{Semi-sup.} 
& CDAC & 77.57 & 67.35 & 74.93 & 80.04 & 61.69 & 69.01 & 29.69 & 8.00 & 23.97 \\
& CDAC-KM & 87.96 & 85.11 & 93.03 & 93.42 & 87.55 & 89.77 & 67.71 & 45.65 & 71.49 \\
& CDAC+ & \textbf{89.30} & \textbf{86.82} & \textbf{93.63} & \textbf{94.74} & \textbf{89.41} & \textbf{91.66} & \textbf{69.84} & \textbf{52.59} & \textbf{73.48} \\
\bottomrule
\end{tabular}
\caption{ \label{results-aux} The clustering results of CDAC+ and its variant methods.}
\end{table*}


\subsubsection{SNIPS} It is a personal voice assistant dataset which contains 14484 utterances with 7 types of intents.

\subsubsection{DBPedia \cite{zhang2015text}} It contains 14 non-overlapping classes of ontology selected from DBPedia 2015 \cite{lehmann2015dbpedia}. We follow \citet{wang-etal-2016-semi} and randomly select 1,000 samples for each classes.

\subsubsection{StackOverflow}
Originally released on Kaggle.com, it contains 3,370,528 title of technical questions across 20 different classes. We use the dataset processed by  \citet{xu-etal-2015-short} who randomly select 1,000 samples for each classes.

\subsection{Baselines}
We compare our method with both unsupervised and semi-supervised clustering methods. 
\subsubsection{Unsupervised} We compare our method with K-means (KM) \cite{macqueen1967some}, agglomerative clustering (AG) \cite{gowda1978agglomerative}, SAE-KM and DEC \cite{xie2016unsupervised} , DCN \cite{yang2017towards} and DAC \cite{chang2017deep}. For KM and AG, we encode the sentence as a 300-dimensional embedding by averaging the pre-trained GloVe \cite{pennington2014glove} word embeddings. We also run K-means on sentences encoded with averaging embeddings of all output tokens of the last hidden layer of the pre-trained BERT (BERT-KM).

\subsubsection{Semi-unsupervised} For semi-unsupervised methods, we compare with PCK-means \cite{basu2004active} , BERT-Semi \cite{wang-etal-2016-semi} and BERT-KCL \cite{hsu2018learning}. For a fair comparison, we change the backbone network of these methods to the same BERT model as ours.

\subsection{Evaluation Metrics}
We follow previous studies and choose three metrics that are widely used to evaluate clustering results: Normalized Mutual Information (NMI), Adjusted Rand Index (ARI), and clustering accuracy (ACC). To calculate clustering accuracy, we use the Hungarian algorithm \citep{kuhn1955hungarian} to find the best alignment between the predicted cluster label and the ground-truth label. All metrics range from 0 to 1. The higher the score, the better the clustering performance.

\subsection{Experimental Settings}
For each run of experiments, we randomly select 25\% of classes as unknown and 10\% of training data as labeled. We set the number of clusters as the ground-truth. Besides, we divide all dataset into training, validation, and test sets. First, we train the model by limited labeled data (containing known intents) and unlabeled data (containing all intents) in the training set. Second, we tune the model on the validation set, which only contains known intents. Finally, we evaluate the results on the test set. We report the average performance of each algorithm over ten runs.

We build our model on top of the pre-trained BERT model (base-uncased, with 12-layer transformer) implemented in PyTorch \cite{Wolf2019HuggingFacesTS} and adopt most of its hyper-parameter settings. To speed up the training process and avoid over-fitting, we freeze all the parameters of BERT except the last transformer layer. The training batch size is 256, and the learning rate is $5e^{-5}$. We use the same dynamic thresholds as DAC \cite{chang2017deep} and set $u(\lambda) = 0.95 - \lambda$, $l(\lambda) = 0.455 + 0.1 \cdot \lambda$, and $\eta = 0.009$.

During the refinement stage, we perform K-means on intent representation $I$ to obtain the initial cluster centroids $U$ and set the stop criteria $\delta_{label}$ as 0.1\%.

\begin{figure*}[!t]
  \centering  
  \includegraphics[width=2\columnwidth ]{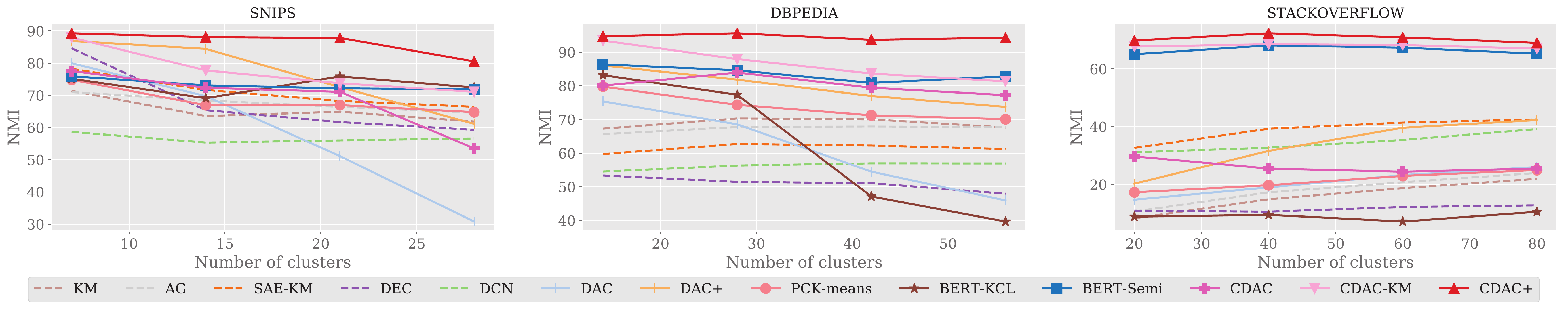}
  \caption{\label{fraction} Influence of the number of clusters on three datasets.}
\end{figure*}

\begin{figure*}[!t]
  \centering  
  \includegraphics[width=2\columnwidth ]{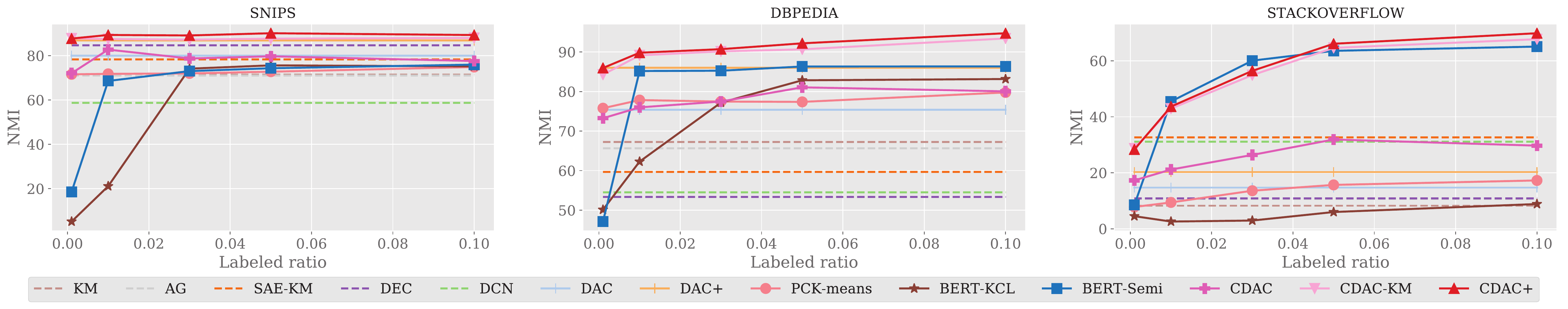}
  \caption{\label{labeled} Influence of the labeled ratio on three datasets.}
\end{figure*}

\begin{figure*}[!t]
  \centering  
  \includegraphics[width=2\columnwidth ]{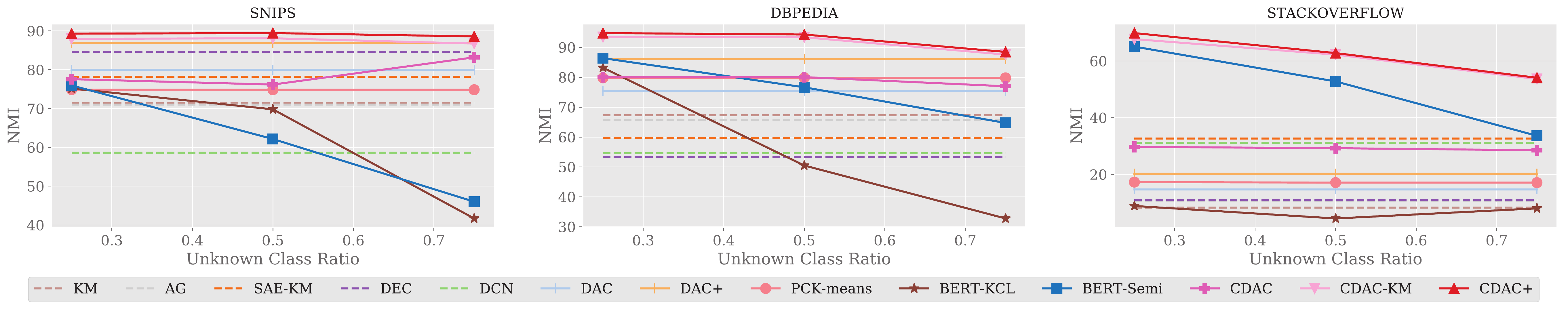}
  \caption{\label{unknown} Influence of the unknown class ratio on three datasets.}
\end{figure*}

\subsection{Results and Discussion}
The results are shown in Table \ref{results-main}. The proposed CDAC+ method outperforms other baselines by a significant margin in all datasets and evaluation metrics. It shows that our method effectively groups sentences based on the intent representations learned with pairwise classification and constraints, and even can generalize to new intents that we do not know in advance. 

The performance of unsupervised methods is particularly poor on DBPedia and StackOverflow, which may be related to the number of intents and the difficulty of the dataset. Semi-supervised methods are not necessarily better than unsupervised methods. If the constraints are not used correctly, it can not only lead to overfitting but also fail to group new intents into clusters. 

Among these baselines, BERT-KM performed the worst, even worse than running K-means on sentences encoded with Glove. Our results suggest that fine-tuning is necessary for BERT to perform downstream tasks. Next, we will discuss the robustness and effectiveness of the proposed method from different aspects.

\subsubsection{Ablation study}
To investigate the contribution of constraints and cluster refinement, we compare CDAC+ with its variant methods, such as performing K-means clustering with representation learned by DAC (DAC-KM) or CDAC (CDAC-KM), CDAC+ without constraints (DAC+), and CDAC+ without cluster refinement (CDAC). The results are shown in Table \ref{results-aux}. 

Most methods have better performance when constraints are added. Compared with DAC+ on StackOverflow, CDAC+ can even increase clustering accuracy by up to 50\%. It shows the effectiveness of constraints. For cluster refinement, DAC+ and CDAC+ consistently perform better than DAC-KM and CDAC-KM. DAC+ even outperforms other baselines on SNIPS and DBPedia. It implies that learning representation only through DAC or CDAC is not enough, and cluster refinement is necessary to get better results.

\begin{figure*}[t]
\begin{subfigure}{.65\columnwidth}
    \centering
    \includegraphics[width=\columnwidth]{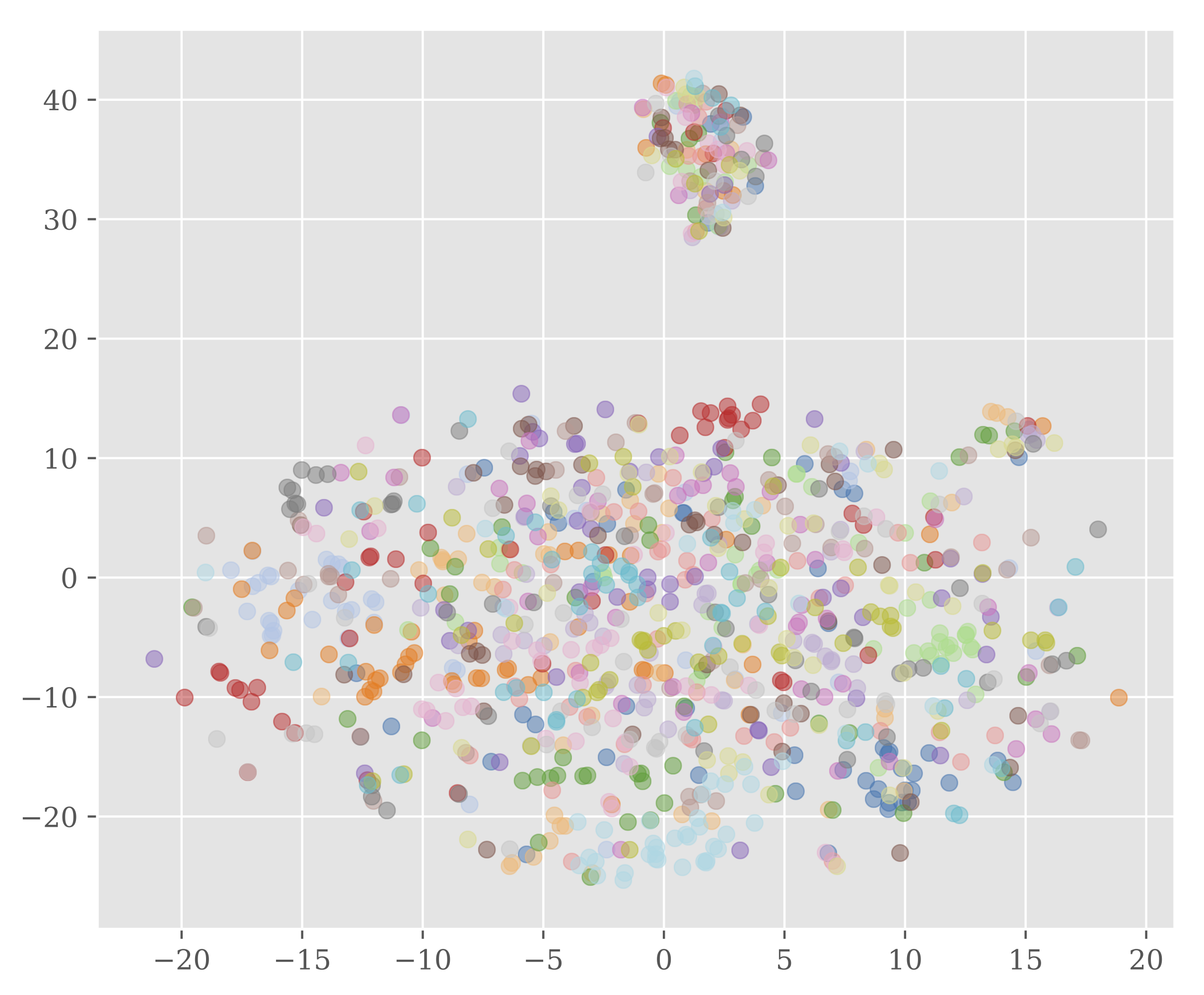}
    \caption{DAC}
\end{subfigure}
\begin{subfigure}{.65\columnwidth}
    \centering
    \includegraphics[width=\columnwidth]{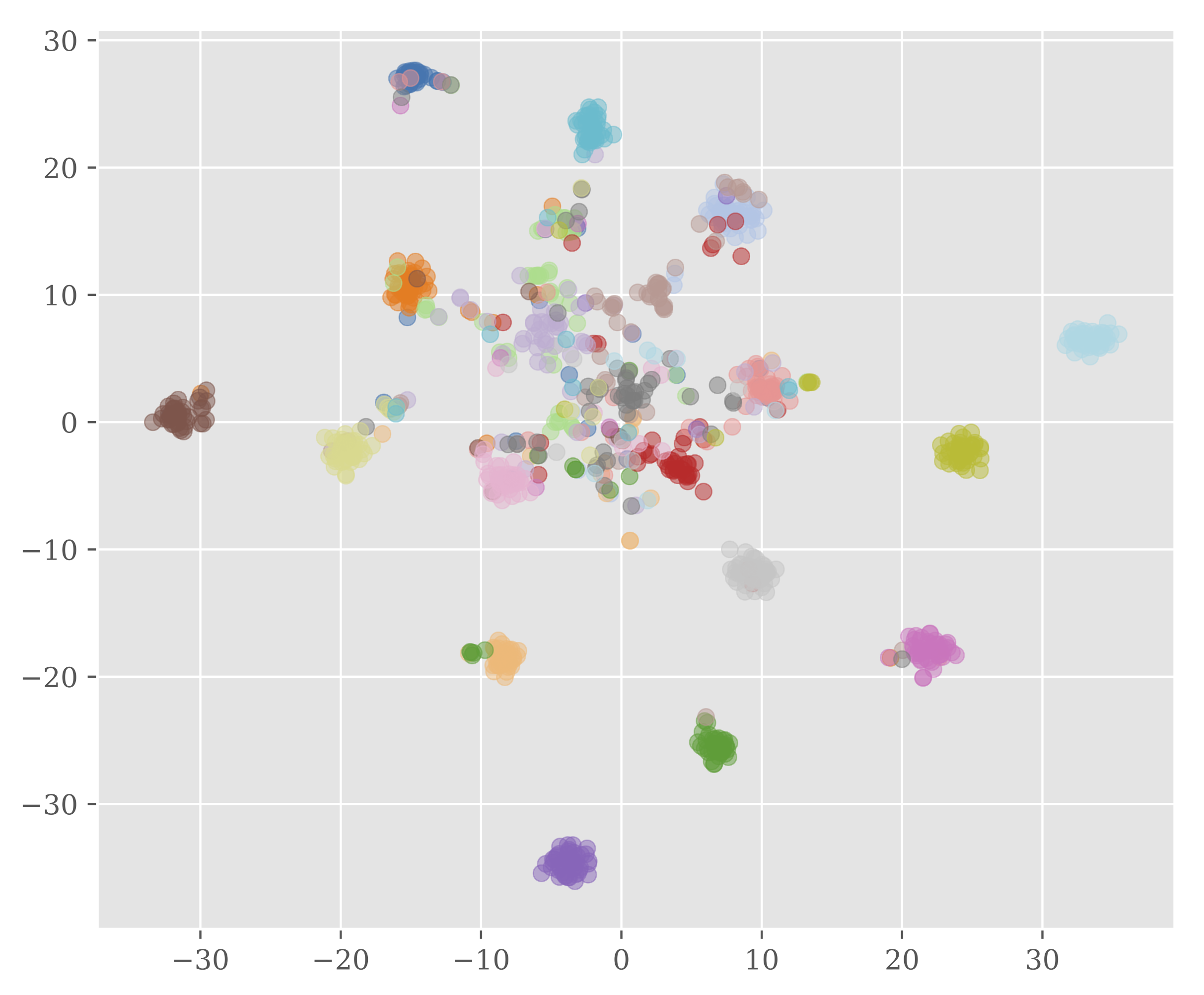}
    \caption{BERT-Semi}
\end{subfigure}
\begin{subfigure}{.75\columnwidth}
    \centering
    \includegraphics[width=\columnwidth]{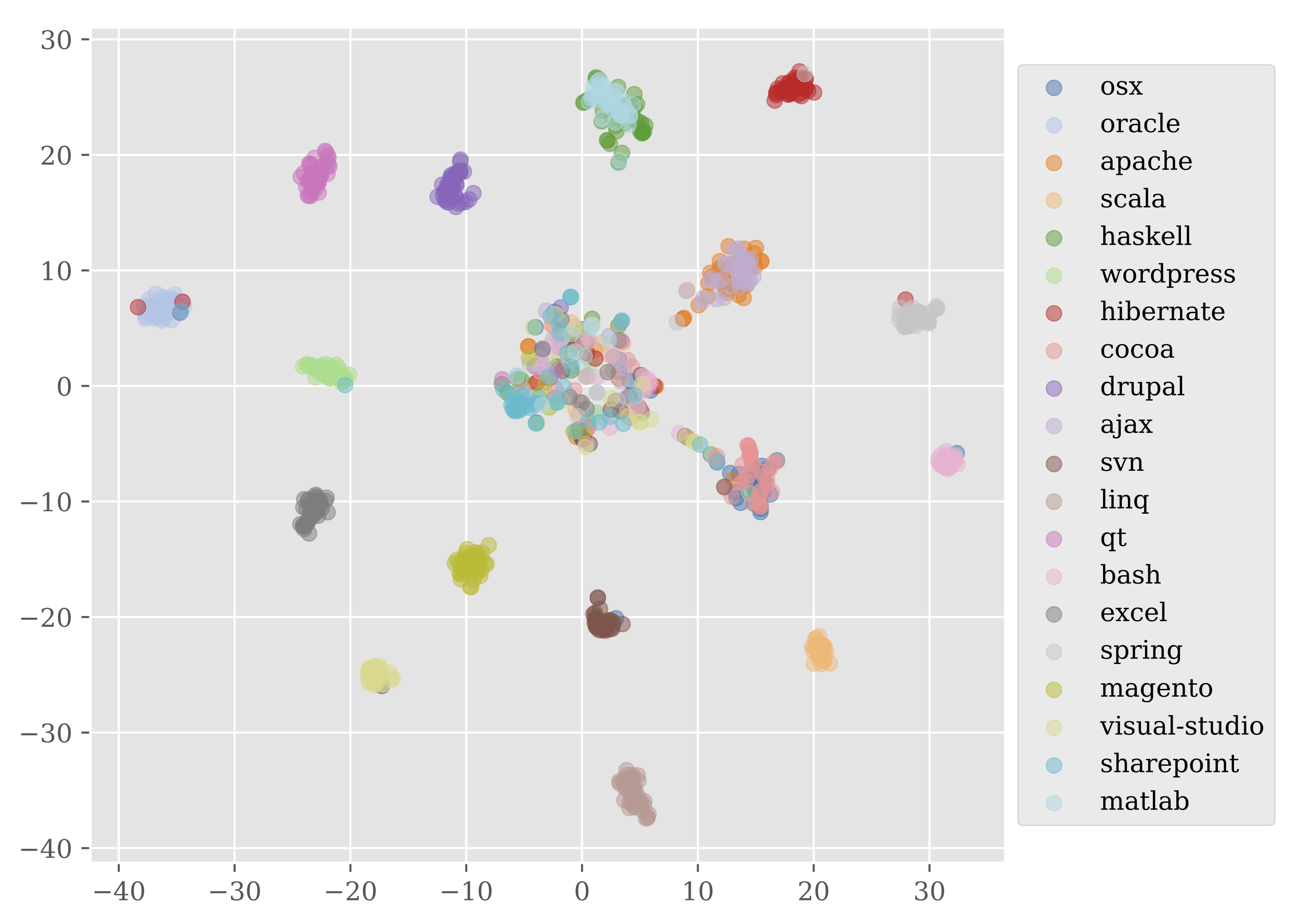}
    \caption{CDAC+}
\end{subfigure}
\caption{Visualization of intent representation learned on StackOverflow dataset.}
\label{tsne2}
\end{figure*}

\begin{figure}[t!]
  \includegraphics[width=0.95\columnwidth ]{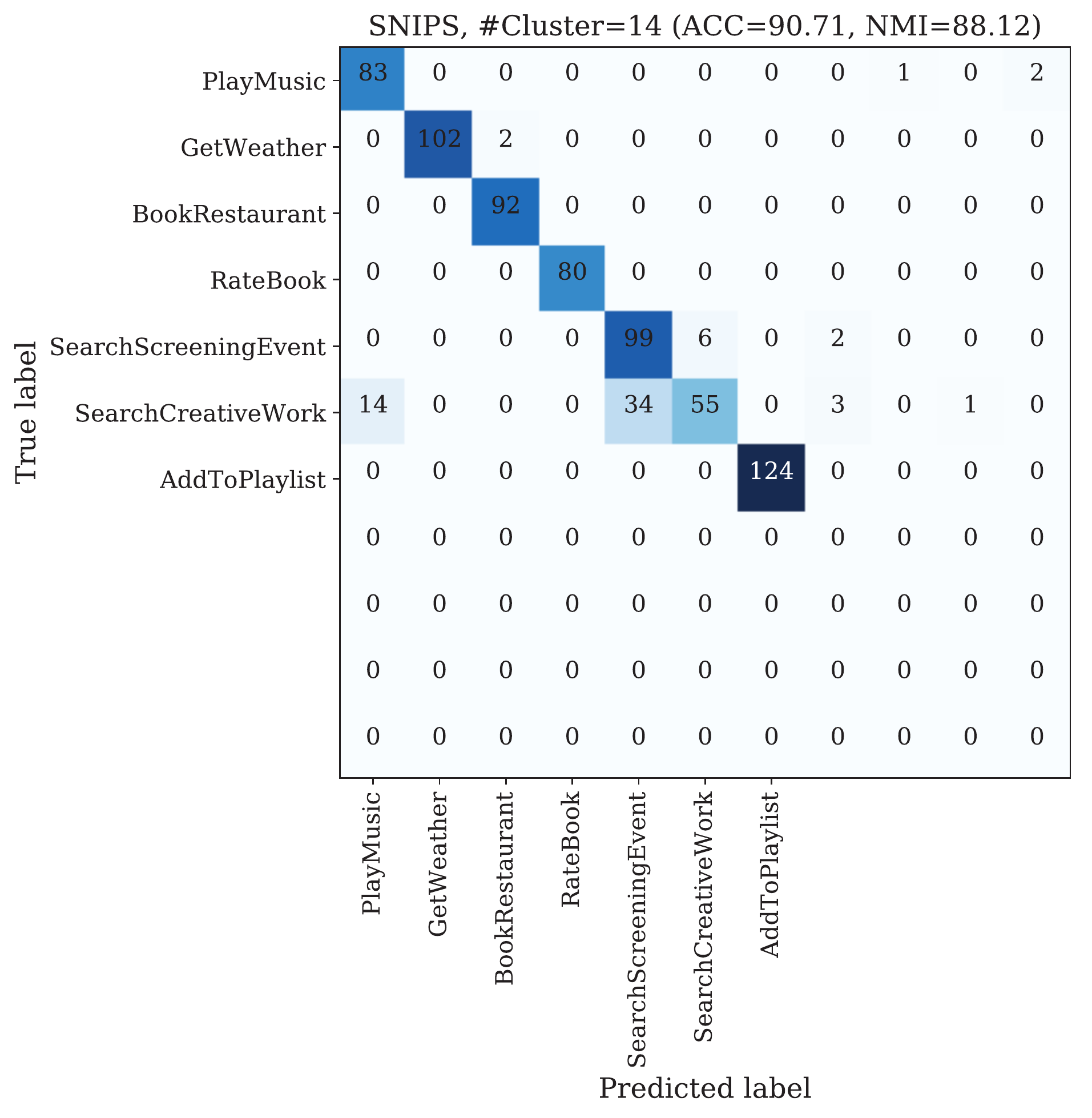}
\caption{ \label{mat} 
Confusion matrix for the clustering results of CDAC+ on SNIPS datasets. The predefined number of clusters is twice of its ground truth. The values along the diagonal represent how many samples are correctly classified into the corresponding class. The larger the number, the deeper the color. We hide empty clusters for better visualization.}
\end{figure}

\subsubsection{Effect of Number of Clusters}
To study whether our method is sensitive to the number of clusters or not, we increase the number of predefined clusters from its ground truth number to four times of it. The results are shown in Figure \ref{fraction}. As the number of clusters increases, the performance of almost all methods except CDAC+ drops dramatically. Besides, our method consistently performs better than CDAC-KM, which demonstrates the robustness of cluster refinement. In Figure \ref{mat}, we use the confusion matrix to analyze the results further. It shows that our method not only maintains excellent performance but is also insensitive to the number of clusters. 

\subsubsection{Effect of Labeled Data}
We vary the ratio of labeled data in training set in the range of 0.001, 0.01, 0.03, 0.05 and 0.1, and show the results in Figure \ref{labeled}. First, even if the ratio of labeled data is much lower than 0.1, CDAC+ still performs better than most baselines. Second, the performance changes the most on the StackOverflow dataset. The reason is that the taxonomy of it can be divided by technical subjects or question types (e.g., what, how, why). It requires the labeled data as prior knowledge to guide the clustering process. The unsupervised methods fail since there is no prior knowledge to guide the clustering process. 

Finally, the NMI score of BERT-Semi is slightly better than CDAC+ when the labeled ratio is 0.01 and 0.03 on StackOverflow. The reason is that BERT-Semi uses instance-level constraints as prior knowledge. It can easily group known intents but fail to group the unknown intents into new clusters. We will discuss it in the next paragraph.

\subsubsection{Effect of Unknown Classes}
We vary the ratio of unknown classes in training set in the range of 0.25, 0.5 and 0.75, and show the results in Figure \ref{unknown}. The higher the ratio of unknown classes, the more new intent classes in the training set. Our method is still robust compared with baselines. In this case, the performance of BERT-Semi drops dramatically. The instance-level constraints they use will cause overfitting and will not be able to group new intents into clusters.

\begin{figure}[t!]
  \centering  
  \includegraphics[width=0.95\columnwidth ]{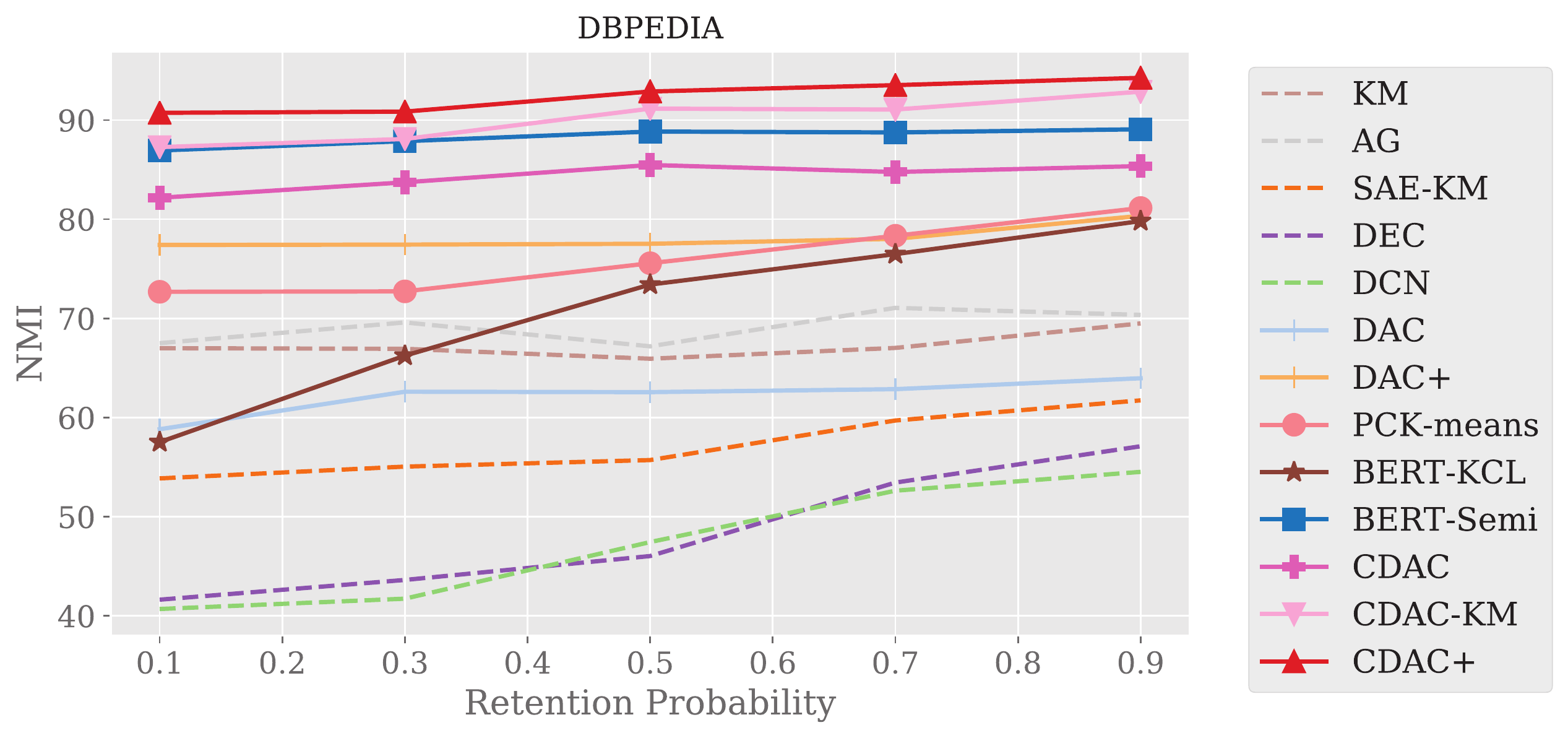}
  \caption{\label{retention_probability} Influence of imbalanced subset on StackOverflow.}
\end{figure}

\subsubsection{Performance on Imbalanced Dataset}
We follow previous works \cite{chang2017deep} and randomly sample subsets of datasets with different minimum retention probability $\gamma$. Given a dataset with N-classes, samples of class 1 will be kept with probability $\gamma$ and class N with probability 1. The lower the $\gamma$, the more imbalanced the dataset is. The results are shown in Figure \ref{unknown}. Our method is not only robust to imbalanced classes but also outperform other baselines trained with balanced classes. The performance of other baselines drops around 3\% to 10\% under different $\gamma$.

\subsubsection{Error Analysis}
We further analyze whether CDAC+ can discover new intents on the test set. In Figure \ref{mat}, we set \textit{BookRestaurant} and \textit{SearchCreativeWork} as unknown in training set. Our method is  still able to find out these intents. Note that some samples of \textit{SearchCreativeWork} are incorrectly assigned to cluster of \textit{SearchScrrenEvent} since they are semantically similar. 

In Figure \ref{tsne2}, we use the t-SNE \cite{maaten2008visualizing} to visualize the intent representation. Compared with other methods, the representation learned by CDAC+ is compact within the class and separable between classes. It shows that our method does learn cluster-friendly representations.

\section{Conclusion and Future Work}
In this paper, we propose an end-to-end clustering method that uses limited labeled data to guide the clustering process for discovering new intents and further refine the cluster results by forcing the model to learn from the high confidence assignments. Extensive experiments show that our method not only yields significant improvements compared with strong baselines but is also insensitive to the number of clusters. In the future, we will try to combine different kinds of prior knowledge to guide the clustering process.

\section{Acknowledgments}
This paper is funded by the National Natural Science Foundation of China (Grant No: 61673235) and National Key R\&D Program Projects of China (Grant No: 2018YFC1707605).

\small \bibliography{aaai20}
\bibliographystyle{aaai}

\end{document}